\documentclass[letterpaper, 10 pt, conference]{ieeeconf}  

\IEEEoverridecommandlockouts                              

\usepackage{cite}
\usepackage{amsmath,amssymb,amsfonts}
\usepackage{algorithmic}
\usepackage{graphicx}
\usepackage{textcomp}
\usepackage{xcolor}
\usepackage{orcidlink}
\usepackage{pifont}   
\usepackage{cite}
\usepackage{censor} 
\usepackage{booktabs}
\usepackage[normalem]{ulem}

\def\BibTeX{{\rm B\kern-.05em{\sc i\kern-.025em b}\kern-.08em
    T\kern-.1667em\lower.7ex\hbox{E}\kern-.125emX}}
\title{LLM-Guided Compositional Continual Learning for Robot Manipulation
with Gated Low-Rank Experts}

\author{Nilay Kushawaha\textsuperscript{*1}, Abdalla Swikir\textsuperscript{2}, and Fares Abu-Dakka\textsuperscript{1}
\thanks{\textsuperscript{*}Corresponding author.
\textsuperscript{1}Mechanical Engineering Program, New York University Abu Dhabi, Abu Dhabi, UAE.
\textsuperscript{2}Mohammed bin Zayed University for Artificial Intelligence, Abu Dhabi, UAE.}
}
\begin{document}
\maketitle

\begin{abstract}
General-purpose manipulation robots must acquire new skills over time without
forgetting prior ones or retraining from scratch. We present a continual
learning framework that represents each skill as a dedicated expert sub-network in a
Progressive Neural Network (PNN). New tasks reuse prior knowledge through low-rank adaptation (LoRA) modules on lateral connections controlled by learnable per-parent gates, enabling explicit attenuation of individual transfer paths while reducing the parameter cost of dense PNN laterals. For long-horizon deployment, a frozen
local large language model decomposes natural-language instructions into
sub-tasks, routes them to suitable experts, and requests a new expert only when
the library is insufficient. Using a shared task-space action interface, we evaluate the framework on twelve MetaWorld tasks and in a Sawyer-to-Panda-to-UR5e cross-embodiment transfer experiment, showing improved task success relative to learning from scratch when demonstrations are limited. Real-robot manipulation experiments further demonstrate that the proposed framework supports compositional execution and expert orchestration.
\end{abstract}


\section{Introduction}

Robots operating in unstructured environments must acquire new manipulation
skills throughout their lifetime without losing previously learned ones. This
requirement is difficult to satisfy with the dominant paradigm of training a
single monolithic policy across all tasks. When new tasks are added, previously
acquired knowledge can be partially or completely overwritten, leading to
catastrophic forgetting~\cite{ewc}. Continual learning (CL) addresses this problem by
enabling incremental skill acquisition while preserving earlier capabilities.
However, forgetting is not the only limitation of monolithic policies \cite{kushawaha2024synapnet}. Because
different skills are entangled within the same set of parameters, it is difficult
to identify which part of the network is responsible for a particular behavior or
to quantify how much a new skill reuses prior knowledge. In safety-critical
robotics~\cite{binny2026safe}, this lack of transparency limits inspection,
verification, and selective deployment of individual competencies.

\begin{figure}
	\centering
	\includegraphics[width=\columnwidth]{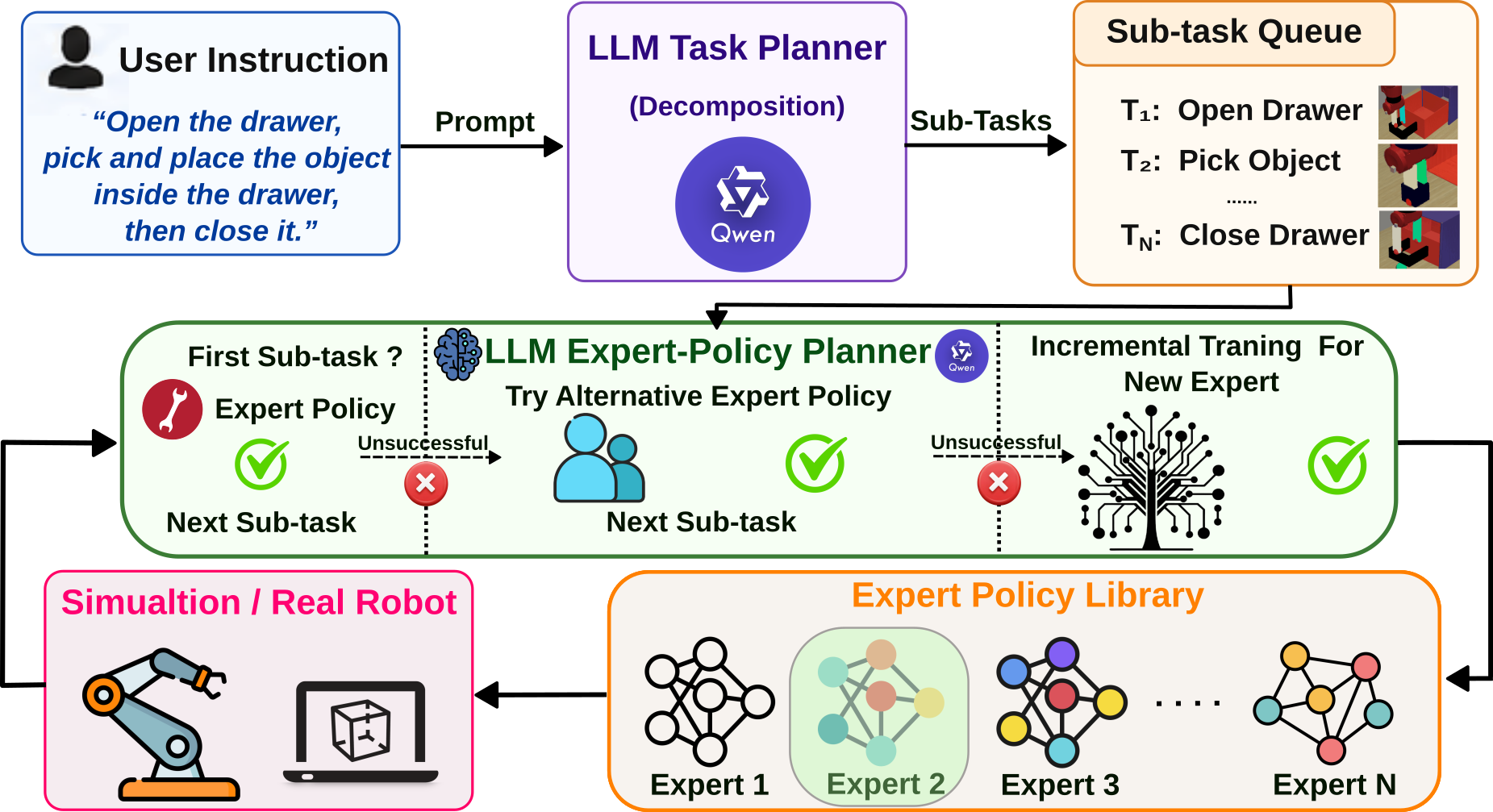}
	\caption{A frozen LLM decomposes a long-horizon instruction into sub-tasks and
		routes each to an expert in a continually growing library of frozen PNN columns, training a new expert policy for unseen tasks.}
	\label{fig:abs_achitecture}
\end{figure}

We address these issues by learning each skill as a dedicated expert that is
preserved after training and composed only when needed, as shown in Fig. \ref{fig:abs_achitecture}. Our architecture is based
on a Progressive Neural Network (PNN)~\cite{pnn}, where every task is assigned a
new column, or sub-network, that is frozen once learned. This prevents parameter overwriting in previously learned columns. To enable forward knowledge transfer with a lower per-connection parameter cost than dense lateral connections, prior knowledge is reused through LoRA-style low-rank lateral adapters ~\cite{lora}. Each adapter is modulated by a learnable
per-parent gate, allowing the new column to draw selectively from relevant prior
experts while suppressing unhelpful ones. This design reduces the quadratic
parameter growth of vanilla PNNs, mitigates negative knowledge transfer, and exposes
transfer as a measurable quantity. Depending on the experiment, experts are trained either with soft actor-critic (SAC) augmented by behavior cloning or from demonstrations using conditional flow matching ~\cite{sac}.

A library of reusable experts becomes most useful when paired with a planner that
can decide which skill to invoke. We therefore integrate a frozen, locally served
large language model (LLM) as a high-level supervisor. Given a natural-language
instruction, the LLM decomposes it into an ordered sequence of sub-tasks, selects
the appropriate expert for each step, and requests that a new expert be trained
only when the current library is insufficient. The LLM operates purely at the
symbolic planning level and is never fine-tuned, keeping task reasoning separate
from low-level control and making the high-level task decomposition and routing explicit.

Using shared task-space action conventions, we also evaluate the gated transfer mechanism across robot embodiments. We study transfer from a Sawyer arm in MetaWorld~\cite{mclean2025metaworld}
to Panda and UR5e arms in Robosuite~\cite{robosuite2020}, and finally to a real Franka Research~3 (FR3) arm performing a long-horizon manipulation task. In the simulated cross-embodiment experiments, the framework achieves higher mean task success than learning from scratch at every tested demonstration budget. The main contributions of this article are summarized as follows:

\begin{itemize}
\item We introduce a continual reinforcement learning (CRL) framework that combines frozen PNN
columns with LoRA lateral adapters and learnable per-parent gates, enabling selective and parameter-efficient forward transfer.
\item We develop a frozen local LLM planner that decomposes long-horizon instructions,
routes sub-tasks to learned expert policies, and requests a new expert when the
library cannot solve the task, while remaining separate from low-level control.
\item We evaluate the framework against representative CL methods in the MetaWorld simulator, and demonstrate cross-embodiment transfer to the Robosuite simulator and then to an FR3 robot.
\item We demonstrate a six-stage real-world manipulation sequence on the FR3, where the LLM planner coordinates dedicated expert policies across all the sub-tasks.
\end{itemize}

\begin{figure*}[t]
	\centering
	\includegraphics[width=\textwidth]{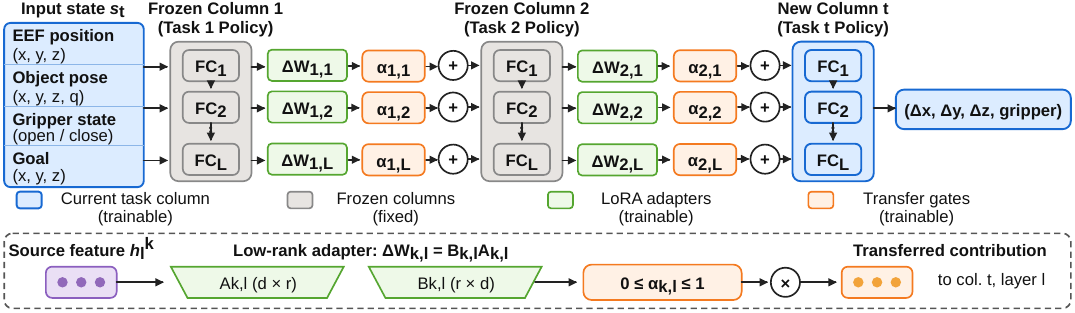}
	\caption{Our proposed PNN-LoRA architecture. Each previously learned task is represented by a frozen column, and a new trainable column
		receives forward transfer from all prior columns through low-rank lateral
		adapters.}
	\label{fig:lora_architec}
\end{figure*}

\section{Related Work}
\label{sec:related}

\subsection{Continual Learning for Robot Manipulation}

CL methods are commonly grouped into three categories:
\emph{Regularization} methods preserve previous knowledge by penalizing changes
to parameters that are important for past tasks~\cite{ewc}. \emph{Rehearsal}
methods revisit stored samples from earlier tasks, or generate synthetic ones,
to reduce forgetting~\cite{er}. \emph{Architectural} methods allocate separate
capacity to new tasks, either by expanding the network with task-specific
modules~\cite{den,pnn} or by partitioning a fixed-size model through pruning or
masking~\cite{packnet}.

In robotic manipulation, CL is often studied with reinforcement learning
(RL) or imitation learning (IL). Li \emph{et al.}~\cite{li2024continual} study
dexterous manipulation in simulation using an offline CRL student--teacher
framework with replay. LEGION~\cite{legion} accumulates knowledge from a
continuous task stream through a Bayesian nonparametric knowledge space, a SAC
policy, and experience replay, while using language embeddings to infer tasks in
long-horizon settings. CLARE~\cite{clare} adds encoder-decoder adapters to a
frozen flow-matching vision-language-action (VLA) model, expanding only when a
feature-novelty score indicates a new task and routes without task IDs.
SPECI~\cite{speci} builds a self-evolving skill codebook with frozen prior
entries and decomposes shared and task-specific parameters. COLADA~\cite{colada}
freezes an action-chunking transformer backbone, learns one LoRA adapter per
skill from teleoperated demonstrations, and acquires missing skills through
grounded human \emph{dialogue}.

While these methods are effective, many rely on large frozen VLA backbones that
are costly to train, store, and deploy. In contrast, we use a PNN composed of
compact task-specific expert policies, connected through LoRA-style low-rank lateral adapters~\cite{lora}. This design emphasizes selective forward transfer across both tasks
and robot embodiments, a setting not addressed by the above methods.



\subsection{LLMs as High-Level Planners for Long-Horizon Manipulation}

Recent work increasingly uses LLMs to decompose
natural-language commands into sub-goals rather than to generate low-level robot
actions directly. Huang \emph{et al.}~\cite{huang2022language} prompt a frozen
LLM to map household goals to mid-level actions from a fixed vocabulary.
LEAGUE++~\cite{leaguepp} combines task-and-motion planning with SAC, using an
LLM to produce symbolic operators and design dense rewards, while
warm-starting each new policy from the embedding-nearest prior skill.
CLIMB~\cite{climb} constructs a symbolic PDDL world model and introduces new
predicates after execution failures. LRLL~\cite{lrll} grows a library of
LLM-written \emph{code} skills through a gradient-free wake--sleep procedure.
Tan~\emph{et al.}~\cite{mmllm_longhorizon} decompose long-horizon tasks over a
tree-structured skill library with GPT-4V perception, querying a human when a
skill is unavailable. LCB~\cite{lcb} represents sub-goals as learnable latent
tokens, but fine-tunes the LLM end-to-end and couples it to a single monolithic
policy. These approaches similarly use LLMs primarily for high-level planning rather than direct control. Our framework differs by using a frozen local LLM
to decompose tasks, select among learned expert policies, and request that a new expert be trained when no suitable expert is available.

\section{Problem Statement}
\label{sec:problem}

We study lifelong manipulation, where a robot must acquire skills sequentially,
reuse prior knowledge across tasks and embodiments, and compose learned skills
from language. We separate this setting into two layers: a low-level control
layer, in which each skill is represented by a state-feedback policy, and a
high-level planning layer, in which language selects and orders those skills.
Language is therefore used only for planning and never as an input to the control
policy. This formulation leads to three requirements.

\textbf{(P1) Incremental skill acquisition.} Skills arrive sequentially as tasks
$\{\mathcal{T}_n\}_{n=1}^{N}$, with $N$ unknown in advance. Each task is an MDP
$\mathcal{T}_n=(\mathcal{S},\mathcal{A},P_n,r_n,\rho_n^0,\gamma)$, sharing the
state space $\mathcal{S}\subseteq\mathbb{R}^{d_s}$ and task-space action space
$\mathcal{A}\subseteq\mathbb{R}^{d_a}$ with all other tasks. The action is an
end-effector command $[\Delta x,\Delta y,\Delta z,g]$, while the dynamics
$P_n$, reward $r_n$, and initial distribution $\rho_n^0$ may differ across
tasks. At stage $n$, the learner can interact only with $\mathcal{T}_n$ and has
access to a small demonstration set $\mathcal{D}_n$; previous environments and
data are unavailable. For each task, we learn a dedicated expert policy $\pi_n$
that maximizes
\begin{equation}
J_n=\mathbb{E}_{\rho_n^0,\,\pi_n,\,P_n}\Big[\textstyle\sum_{t\ge0}\gamma^t\, r_n(s_t,a_t)\Big].
\label{eq:return}
\end{equation}
The lifelong learner must satisfy two constraints: \textbf{(R1) no forgetting},
so learning a new task does not degrade previously learned experts, and \textbf{(R2)
selective forward transfer}, so useful prior experts improve learning while
unhelpful lateral connections are suppressed.

\textbf{(P2) Compositional orchestration.} The learned experts form a library
$\Pi=\{(\pi_i,m_i)\}$, where each policy $\pi_i$ is paired with a language
descriptor $m_i$. Given a long-horizon instruction $L$ that cannot be solved by a
single policy, the planner decomposes it into sub-task descriptions and assigns
each one to an expert,
\begin{equation}
(\hat{m}_1,\dots,\hat{m}_K)=\Phi(L),\qquad e_k=\rho(\hat{m}_k,\Pi).
\label{eq:route}
\end{equation}
Here, $\Phi$ returns an ordered list of $K$ sub-tasks, and $\rho$ routes each
$\hat{m}_k$ to a matching expert in the library. If no expert applies,
$e_k=\varnothing$, triggering acquisition of a new expert under (P1). The
selected expert is then executed closed-loop on the robot state. We require both
$\Phi$ and $\rho$ to be implemented by a frozen, general-purpose planner rather
than a task-specific trained module.

\textbf{(P3) Cross-embodiment reuse.} The same library should support transfer
across robot embodiments. Given experts learned on a source robot, such as a
Sawyer in MetaWorld or a Panda in Robosuite, the learner should acquire a new
skill on a target embodiment, such as a UR5e or a real FR3, from
only a few target demonstrations. Since all policies share the same task-space
action interface, prior experts can be reused when learning downstream skills on
new robots.

\section{Methodology}
\subsection{PNN-LoRA: Gated Low-Rank Lateral Transfer}
\label{sec:pnnlora}
A PNN represents each task with a dedicated network
column whose parameters are frozen after training~\cite{pnn}. When task $n$
arrives, a new column is appended while all previous columns $1,\dots,n{-}1$
remain fixed as shown in Fig. \ref{fig:lora_architec}. Transfer to the new task occurs only through lateral connections
from these frozen columns. Each column consists of blocks of width $d$. Block
$\ell$ has its own affine map $f^n_\ell$ and, for $\ell\ge2$, receives one
lateral adapter $L^n_{\ell,j}$ from every previous column $j<n$. With
$h^n_0=s$, the activations are
\begin{equation}
h^n_\ell=\phi\Big(f^n_\ell(h^n_{\ell-1})
      +\sum_{j<n} g^n_{\ell,j}\,L^n_{\ell,j}\!\big(h^{\,j}_{\ell-1}\big)\Big),
\label{eq:column}
\end{equation}
where $\phi$ is a ReLU nonlinearity for the hidden blocks, $h^{\,j}_{\ell-1}$ are the \emph{frozen}
features of parent column $j$, and $g^n_{\ell,j}\in(0,1)$ is a per-parent gate.
The final block outputs the mean and log-standard-deviation over the action space $\mathcal{A}$, which together give the sampled
action and its associated uncertainty. Because only the new column is trainable,
learning task $n$ cannot alter any earlier policy $\pi_m$ with $m<n$, so every
prior skill is preserved exactly.

In the original PNN each lateral
$L^n_{\ell,j}$ is a dense $d\times d$ matrix that is always active, which has two
drawbacks. First, connecting every new column to all predecessors makes the
lateral parameter count grow as $\Theta(n^2 d^2)$, quickly dominating memory
and training cost. Second, the new column is forced to
absorb features from \emph{all} prior experts, including unrelated ones; these
irrelevant signals act as noise that can slow convergence or push final
performance below that of learning the task in isolation, a failure mode known as
negative transfer.

We address both issues through the design of the lateral $L^n_{\ell,j}$ and the
gate $g^n_{\ell,j}$ in Eq.~\ref{eq:column}. First, following LoRA~\cite{lora}, we
replace each dense lateral with a low-rank factorization
$L^n_{\ell,j}(h)=h\,A^n_{\ell,j}B^n_{\ell,j}$, where
$A^n_{\ell,j}\!\in\!\mathbb{R}^{d\times r}$ and
$B^n_{\ell,j}\!\in\!\mathbb{R}^{r\times d}$ with rank $r\ll d$. This lowers the
per-lateral parameter count from $d^2$ to $2rd$ and confines transfer to an
$r$-dimensional subspace of the parent's features, regularizing it when
target-task data is scarce. Second, we attach a learnable gate to every lateral,
$g^n_{\ell,j}=\sigma(\alpha^n_{\ell,j})$, with $\sigma$ the logistic function and
$\alpha^n_{\ell,j}$ a free scalar, so the signal expert $j$ contributes to block
$\ell$ is the gated map
$T^n_{\ell,j}=\sigma(\alpha^n_{\ell,j})\,A^n_{\ell,j}B^n_{\ell,j}$. Because the
gate multiplies the lateral output and is optimized jointly with the policy, it
is driven toward $1$ when its parent expert lowers the task loss and toward $0$
when it does not. The column therefore engages only the predecessors that benefit
the current task, realizing the selective transfer a dense, always-on PNN cannot
express.

Beyond aiding learning, the gates make transfer \emph{measurable}. Let
$c^n_j=\sum_\ell\|g^n_{\ell,j}L^n_{\ell,j}(h^{\,j}_{\ell-1})\|$ be the magnitude
of the signal expert $j$ injects into column $n$, and
$c^n_0=\sum_\ell\|f^n_\ell(h^n_{\ell-1})\|$ that of the column's own
contribution. The transfer share
\begin{equation}
u^n_{j}=c^n_j\Big/\Big(c^n_0+\textstyle\sum_{k<n}c^n_k\Big)\in[0,1]
\label{eq:transfer}
\end{equation}
is then the fraction of column $n$'s output borrowed from prior expert
$j$, letting us read off which skills each new task reuses.

\subsection{Reinforcement Learning Training}
\label{sec:rl}

Each new column is trained with SAC~\cite{sac}, an off-policy maximum-entropy
actor--critic method, augmented with a behavior-cloning (BC) loss from the task
demonstrations $\mathcal{D}_n$. The actor is the newly added PNN column
$\pi_{\theta_n}$ described in Sec.~\ref{sec:pnnlora}. It is trained with two
critics, $Q_{\phi_1}$ and $Q_{\phi_2}$, and their delayed target networks. After
training, only the actor column is frozen and added to the expert library.

The critics minimize the standard soft Bellman residual using the clipped
double-$Q$ target. The actor is optimized with
\begin{equation}
\begin{split}
\mathcal{L}_{\pi}={}&\mathbb{E}_{\mathcal{B}}\big[\eta\log\pi_{\theta_n}(\tilde a\,|\,s)-\textstyle\min_i Q_{\phi_i}(s,\tilde a)\big]\\
&+\lambda_{\mathrm{BC}}\,\mathbb{E}_{\mathcal{D}_n}\big\|\mu_{\theta_n}(s)-a\big\|^2,
\end{split}
\label{eq:actorloss}
\end{equation}
where $\tilde a\sim\pi_{\theta_n}(\cdot|s)$ is sampled through the
reparameterization trick. The first term is the SAC objective, which maximizes
expected value while encouraging exploration through the entropy temperature
$\eta$, tuned automatically to a target entropy. The second term anchors the
policy to expert demonstrations and provides a dense learning signal when task
rewards are sparse.

Training batches are formed from both online replay and a fixed demonstration
buffer, so the learner continues to use expert states while improving through
interaction. During each task, only the newly added column is updated. The gates $\alpha^n$ use a higher learning rate without weight decay, the LoRA factors $A^n,B^n$ use weight decay $\lambda_L$, and the column-specific layers $f^n$ use the base optimizer configuration. Gradients are norm-clipped for
stability. Once the evaluation success rate remains above a preset threshold,
the column is frozen and the learner proceeds to the next task.

\subsection{LLM-Based Planning}
\label{sec:llm}

At deployment, the expert library is coordinated by a \emph{frozen}, locally
served LLM, Qwen2.5-7B~\cite{qwen25} running through Ollama. The LLM is never
fine-tuned and outputs only structured JSON decisions, not low-level robot
actions. Routing is possible because each expert is stored with a short
natural-language descriptor of its capabilities. For example,
\texttt{box-close-v3} is associated with skills such as \emph{reach},
\emph{grasp handle}, and \emph{pick-place}. Expert selection is therefore
performed in language space over skill descriptors rather than over policy
parameters, and the library persists across sessions so each deployment starts
from all previously acquired experts.

The planner operates in two stages, following Sec.~\ref{sec:problem}.
First, during \emph{decomposition}, it receives a compound instruction $L$ and a
catalog of available environments, then returns an ordered list of sub-goals,
each paired with a candidate skill. Second, during \emph{routing and
supervision}, it selects an expert for the current sub-goal by comparing the
sub-goal description with the library descriptors. Based on the running success
rate, the required threshold, and the remaining switch budget, it issues one of
three decisions: \texttt{keep\_expert}, \texttt{switch\_expert}, or
\texttt{train\_new\_expert}.

Execution proceeds one sub-goal at a time in a long-horizon environment where
each sub-goal is presented in its corresponding scene. The selected expert is
rolled out closed-loop on the robot state for a small number of trials, and the
sub-goal is accepted if it succeeds in the majority of them. If it fails, the LLM
switches to the next candidate expert. When the switch budget is exhausted and no
existing expert reaches the required performance, the framework invokes a
\emph{train-on-failure} fallback in which a new PNN column is added and trained for that
skill, re-evaluated on the sub-goal, and then registered with its descriptor in
the library. Thus, the skill library
grows only when the current experts are insufficient, while planning remains
symbolic and control remains learned.

\subsection{Cross-Embodiment and Real-Robot Transfer}
\label{sec:xe}
All embodiments considered in this work are controlled through the same
task-space interface where each policy outputs an end-effector displacement and a
gripper command $[\Delta x,\Delta y,\Delta z,g]$, while the robot's low-level
controller converts this command into joint motion. This shared interface allows
a column trained on one robot to serve as a lateral parent for a new column on
another embodiment. To learn a skill on a target platform, such as a Robosuite
Panda, UR5e, or real FR3, we append a new column and connect its LoRA laterals
(Eq.~\ref{eq:column}) to the previously frozen experts, including the $12$
MetaWorld columns and any Robosuite columns learned later. During training, the
new column learns which prior experts to reuse and which lateral signals to
suppress, reducing interference from unrelated skills.

Since target demonstrations from teleoperation or real robots are limited, the
new column is trained by imitation using a flow-matching policy~\cite{flowmatching}
from $N$ demonstrations. Instead of directly regressing actions, the policy
learns a state-conditioned velocity field over the action space. At deployment,
an action is generated by sampling from a Gaussian prior and integrating this
velocity field for a fixed number of Euler steps; the resulting endpoint is used
as the commanded $[\Delta x,\Delta y,\Delta z,g]$ (along with a 3D rotation vector for the complete pose experiments), which is further sent to the low-level controller.
\begin{figure}[t]
	\centering
	\includegraphics[width=\columnwidth]{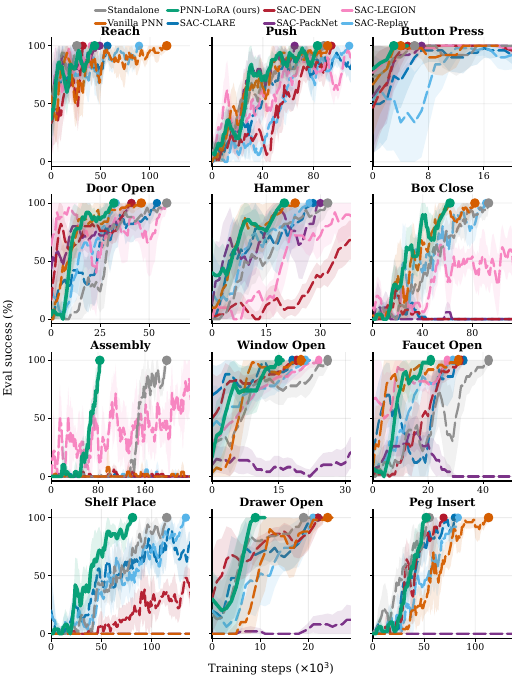}
	\caption{Evaluation success rate versus training steps for each MetaWorld task.
	PNN-LoRA converges faster than the standalone, vanilla-PNN, and
	continual-learning baselines on every task after the first, and learns the
	hard tasks where most baselines fail.}
	\label{fig:train_succ_vs_steps_all}
\end{figure}
\begin{figure*}
	\centering
	\includegraphics[width=0.95\textwidth]{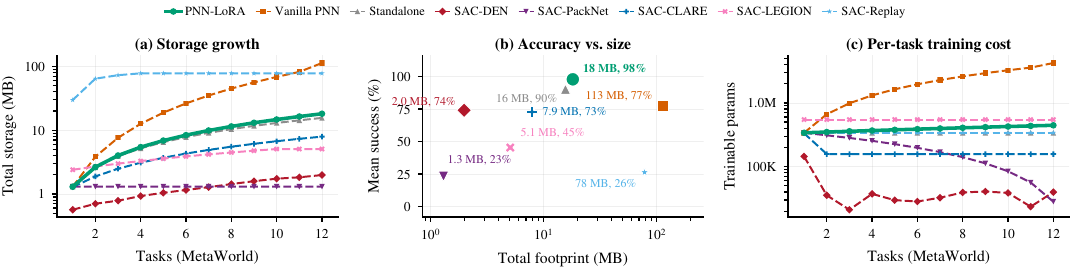}
	\caption{Model size and trainable parameters per task along the sequence.
	PNN-LoRA stays close to the standalone policy, whereas the dense-lateral
	vanilla PNN grows rapidly as more tasks are added.}
	\label{fig:efficiecny_plot}
\end{figure*}
\section{Experiments \& Evaluation}
This section describes the experimental setup and evaluates the proposed
framework in simulation, cross-embodiment transfer, pose-control, and real-world
settings, together with comparisons against benchmark methods.
\subsection{Data Collection}
In the MetaWorld simulation experiments \cite{mclean2025metaworld}, we train on the $12$ tasks listed in
Table~\ref{tab:success_all_methods}.  Each task is
learned using SAC through online environment interaction, while $50$ successful
scripted rollouts are stored in a fixed demonstration buffer and used as an
auxiliary behavior-cloning signal to guide exploration and speed up policy
learning.

For the cross-embodiment study, we use five Robosuite tasks: Panda pick-place
with a \textit{can}, \textit{milk carton}, and \textit{bread}, and UR5e
pick-place with a \textit{milk carton} and \textit{bread}. All tasks share the
same $4$-D operational-space position interface. The Panda \textit{can}
demonstrations are taken from the Robosuite library, while the remaining Panda
demonstrations are collected through keyboard teleoperation. UR5e demonstrations
are recorded directly on the UR5e robot in the Robosuite simulator.

For the real-world experiments, we collect demonstrations on an FR3
arm for empty-glass, hang-glass, box-open, box-close, and pick-place with a bowl and a cup. The robot is
teleoperated using a Force Dimension\texttrademark{} Lambda 7 haptic device to collect the demonstration data. Initial and goal configurations are randomized across demonstrations, and each
object is equipped with an AprilTag so its $6$-DOF pose can be estimated in real
time. We collect $35$ demonstrations per task, using $30$ for training and $5$
for validation.
\subsection{Model Architecture \& Baseline Methods}

Each PNN-LoRA column contains four fully connected blocks. A $39$-D input is
passed through three $400$-unit ReLU layers and mapped to an $8$-D output,
representing the mean and log standard deviation of the action distribution. The
MetaWorld library contains $12$ such columns, trained with the objective in
Eq.~\ref{eq:actorloss}. We use Adam with learning rate $10^{-3}$ for the gates
and $10^{-4}$ for the LoRA factors and column-specific layers, with weight decay
$10^{-3}$ applied to the LoRA factors. Training uses discount factor $0.99$,
batch size $256$ split equally between online and demonstration data, and
$\lambda_{\mathrm{BC}}=2$. The two critics are independent three-layer
$400$-unit networks and are discarded after the corresponding actor column is
trained and frozen.

The Robosuite and real-robot columns use the same architecture but are trained
only from demonstrations using conditional flow matching. Their $39$-D input
contains the robot-object state, the intermediate action sample, and the flow
time. We train these columns with Adam for $300$ epochs using batch size $64$,
and generate actions at deployment by integrating the learned velocity field with
$20$ Euler steps.

We compare PNN-LoRA against SOTA CL baselines:
Experience Replay~\cite{er}, PackNet~\cite{packnet}, DEN~\cite{den}, and the
recent robot-learning methods LEGION~\cite{legion} and CLARE~\cite{clare}. We
also include a vanilla PNN \cite{pnn} with dense always-on laterals and a standalone
per-task policy to bound the comparison. All baselines are implemented in the
same SAC and behavior-cloning framework and evaluated on the same $12$-task
MetaWorld sequence. For the cross-embodiment experiment, the reference baseline is a flow-matching policy (MLP-flow) with the same backbone, trained from scratch without lateral connections.


\begin{figure*}
	\centering
	\includegraphics[width=0.95\textwidth]{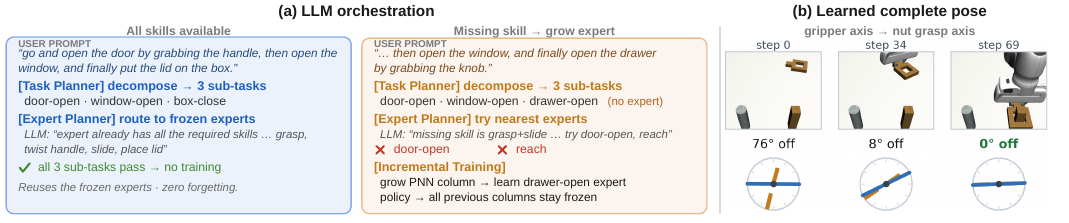}
	\caption{(a) LLM orchestration of a long-horizon task. The instruction is decomposed
	into an ordered set of sub-tasks, each routed to the matching expert; (b) PNN-LoRA policy predicting the actions to handle orientation-aware nut insertion.}
	\label{fig:llm_orchestration_planning}
\end{figure*}

\subsection{MetaWorld Experiments}

A continual learner should use previously acquired skills to learn new ones more
quickly and reliably. We therefore evaluate PNN-LoRA on the $12$-task MetaWorld
sequence, where tasks are learned sequentially and compared against the CL baselines and a standalone policy trained from scratch for every task.

\begin{table}[t]
\centering
\caption{Per-task success rate (\%) on MetaWorld. Mean over 3 seeds; best per task in \textbf{bold}.}
\label{tab:success_all_methods}
\scriptsize
\setlength{\tabcolsep}{1.2pt}
\renewcommand{\arraystretch}{0.85}
\begin{tabular}{@{}lcccccccc@{}}
\toprule
Task & \textbf{Ours} & V-PNN & Standalone & DEN & PackNet & CLARE & LEGION & Replay \\
\midrule
Reach        & 96.7 & 92.3 & 87.3 & \textbf{98.7} & 71.3 & 85.3 & 40.0 & 4.7 \\
Push         & 94.7 & 90.1 & \textbf{96.0} & 86.0 & 2.0 & 40.7 & 90.0 & 0.0 \\
Button       & \textbf{100.0} & 78.0 & 94.2 & \textbf{100.0} & \textbf{100.0} & 88.0 & 0.0 & 0.0 \\
Door         & \textbf{100.0} & 95.1 & 90.7 & 98.7 & 96.0 & 99.3 & 90.0 & 0.0 \\
Hammer       & \textbf{100.0} & 94.7 & 81.3 & 80.7 & 9.3 & 91.3 & 50.0 & 0.0 \\
Box          & \textbf{98.0} & 96.0 & 94.7 & 0.0 & 0.0 & 0.0 & 95.0 & 0.0 \\
Assembly     & \textbf{94.0} & 0.0 & 86.0 & 0.0 & 0.0 & 0.0 & 60.0 & 0.0 \\
Window       & \textbf{100.0} & 97.0 & 73.3 & \textbf{100.0} & 0.0 & 98.7 & 0.0 & 4.7 \\
Faucet       & \textbf{100.0} & \textbf{100.0} & 94.0 & 98.0 & 0.0 & \textbf{100.0} & 0.0 & 88.0 \\
Shelf        & \textbf{91.3} & 0.0 & 90.5 & 34.0 & 0.0 & 72.0 & 0.0 & 54.0 \\
Drawer       & \textbf{100.0} & 97.0 & 98.0 & \textbf{100.0} & 0.0 & \textbf{100.0} & 40.0 & 68.7 \\
Peg          & \textbf{99.3} & 88.7 & 97.3 & 92.7 & 0.0 & 98.7 & 80.0 & 96.7 \\
\midrule
\textbf{Mean} & \textbf{97.8} & 77.4 & 90.3 & 74.1 & 23.2 & 72.8 & 45.4 & 26.4 \\
\bottomrule
\end{tabular}
\vspace{-2mm}
\end{table}

Figure~\ref{fig:train_succ_vs_steps_all} shows evaluation success versus
environment steps for each task. Except for the first task, where no previous
columns are available, PNN-LoRA converges faster than the standalone policy,
vanilla PNN, and the other CL baselines. This can be attributed to the direct effect of positive forward transfer through the gated LoRA laterals, as well as to the selective gating mechanism that preserves useful representations while excluding interfering ones. The effect is most
visible on difficult tasks. In assembly, for example, most baselines fail to
learn, while PNN-LoRA reaches high success and even converges faster than the
standalone policy by reusing earlier manipulation skills.

The faster convergence also translates into strong final performance.
Table~\ref{tab:success_all_methods} reports the final per-task success rates.
PNN-LoRA achieves optimal performance on nearly every task and the highest mean success across the sequence. Several baselines collapse on contact-rich tasks such as box-close and assembly, whereas PNN-LoRA remains stable. It also outperforms vanilla PNN and the standalone policy on most tasks, with the main exceptions being reach and push, where DEN and the standalone policy obtain the best scores.

These gains are achieved with a favorable model footprint. As shown in
Figure~\ref{fig:efficiecny_plot}, PNN-LoRA keeps storage and trainable parameters per task close to the standalone policy while still enabling lateral transfer. This is because low-rank adapters provide transfer at a small fraction of the cost of dense laterals. By contrast, vanilla PNN grows rapidly in both storage and trainable parameters without a corresponding performance gain. Some baselines use fewer parameters, but their success rates are substantially lower, indicating that PNN-LoRA offers a stronger accuracy--efficiency tradeoff.

Finally, Figure \ref{fig:llm_orchestration_planning} illustrates a long-horizon orchestration for a simulation experiment based on the MetaWorld environments. The frozen LLM decomposes a compound instruction into an ordered set of sub-tasks and routes each to the appropriate frozen expert, which then executes it in a closed loop. The example also highlights the train-on-failure fallback. Since no expert was available for the drawer-open task (drawer-open expert policy was withheld for this experiment), the LLM first selected the closest existing policies,
door-open and reach. After both failed, the framework triggered incremental
training to learn a dedicated drawer-open expert.
\subsection{Cross-Embodiment Experiments}

\begin{figure*}[t]
	\centering
	\includegraphics[width=\textwidth]{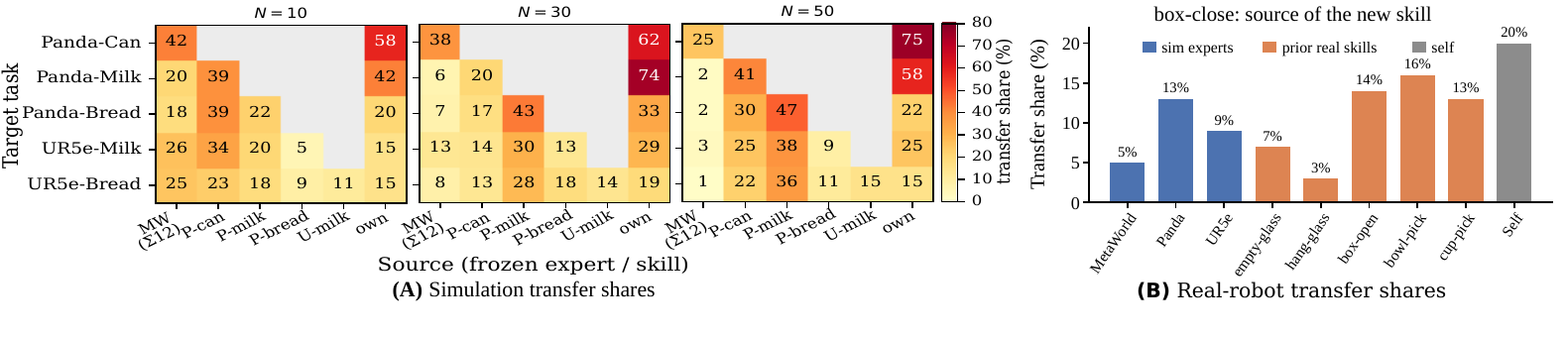}
	\caption{Transfer-share analysis. (A) Per-source shares for Robosuite target
	columns at $N=10,30,50$; the MetaWorld share decreases as data grows, while
	same-family pick-place columns dominate across robot embodiments. (B) Source
	share of the real-robot box-close skill across frozen simulation experts,
	prior real-world skills, and its own path.}
	\label{fig:transfer_share_result}
	\label{fig:box_close_real_world}
\end{figure*}

The cross-embodiment experiments evaluate whether the expert library can transfer
knowledge across both tasks and robot platforms. Figure~\ref{fig:transfer_share_result} (A)
shows the per-source contribution share for each target column at
$N\in\{10,30,50\}$ demonstrations. Two patterns emerge. First, the contribution
from MetaWorld experts decreases as more target data becomes available: from
$42\%$ to $25\%$ on Panda-Can, and from roughly one quarter to at most $3\%$ on the
other targets. This indicates that the coarse cross-simulator prior is most useful
in the low-data regime. As more demonstrations are added, the target column shifts
toward more relevant sources. Since Panda-Can is the first Robosuite task, it has
no same-simulator parent and therefore relies increasingly on its own pathway,
whose share rises from $58\%$ to $75\%$. In contrast, later Robosuite columns
reuse previously learned pick-place experts from the same task family, keeping
their own contribution nearly unchanged, such as $20\%$ to $22\%$ on Panda-Bread
and a constant $15\%$ on UR5e-Bread.

\begin{figure}[t]
	\centering
	\includegraphics[width=\columnwidth]{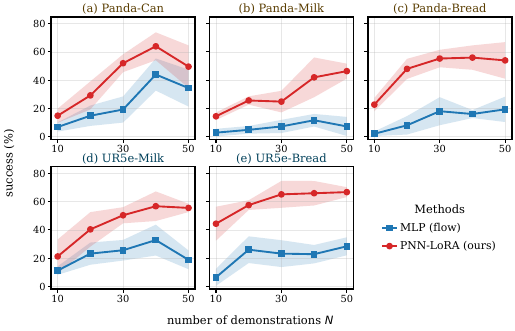}
	\caption{Success rate versus number of demonstrations; PNN-LoRA beats the
	from-scratch MLP-flow baseline at every budget and saturates beyond $N\approx30$.}
	\label{fig:cross_embodiment}
\end{figure}

\begin{figure*}
	\centering
	\includegraphics[width=\textwidth]{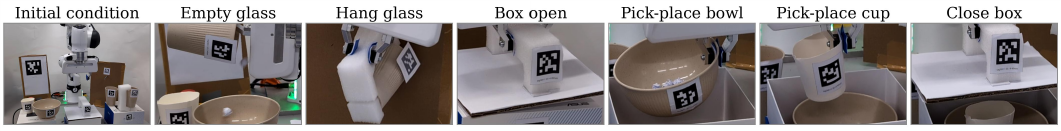}
	\caption{Real-world long-horizon tasks (a) initial state, (b) empty glass, (c) hang glass, (d)  open box,
	(e)--(f) pick-place bowl and cup, and (g) close box. For the task to be
	successful, the policy needs to complete all the sub-tasks.}
	\label{fig:real_world_task}
\end{figure*}


Second, transfer is driven more by task similarity than by robot identity. The
Panda milk and bread columns receive most of their transferred signal from
earlier Panda pick-place experts, particularly can and milk, because these
policies capture the same underlying manipulation pattern. The same trend holds
across embodiments, as the UR5e milk and bread columns also reuse available
Panda and UR5e pick-place experts. This indicates that a skill learned on one
robot can support learning on another when the task structure is shared. The
main exception is UR5e-Bread at $N=10$, where the column relies more heavily on
MetaWorld experts, suggesting that the broader prior can be more
useful than a small number of same-family demonstrations at very low data
budgets.

This selective reuse leads to consistent performance gains. Figure~\ref{fig:cross_embodiment}
compares success rate as a function of the number of demonstrations for PNN-LoRA
and the from-scratch MLP-flow baseline. PNN-LoRA outperforms the baseline on every
task and at every demonstration budget, reflecting the benefit of lateral transfer
that the baseline cannot access. The largest gains appear on bread pick-and-place,
the most challenging task due to the smaller object size, where reusable
pick-place priors are especially valuable. The improvement eventually saturates;
after approximately $30$ demonstrations, the curves flatten, with four of five
tasks changing by at most $5$ points between $N=30$ and $N=50$, and three already
peaking at $N=40$.

\subsection{Complete Pose Control Experiments}
\label{sec:pose}

The preceding experiments used a $4$-D action space containing only
end-effector translation and gripper state. To test whether the proposed
continual mechanism also extends to orientation-aware control, we replace this
interface with Robosuite's $7$-D pose command
$[\Delta p,\Delta r,g]$, where $\Delta r$ is an incremental axis--angle rotation
expressed in the robot base frame. Everything else, including the PNN-LoRA structure, frozen columns, LoRA laterals, and gated transfer mechanism, remains unchanged.

For this experiment, we learn three tasks incrementally from $50$ demonstrations each, along with the online interaction (same as metaworld training): nut
assembly, door opening, and pose-aware cereal pick-place. The resulting policies reach
$84.0\%$, $100.0\%$, and $76.0\%$ success, respectively. Transfer follows the
same pattern observed in the translation-only setting. The door-opening column
strongly gates onto the nut-assembly expert and converges
in $10.6$k frames, compared with $106$k frames for its parent.

Figure~\ref{fig:llm_orchestration_planning}(b) illustrates this behavior. The
square nut is initialized with an arbitrary yaw, while the gripper begins
$76^\circ$ away from the only feasible grasp axis. The policy rotates the wrist
toward that axis, grasps the handle, and carries the nut over the fixed peg,
ending with less than one degree of residual orientation error. A rollout is
counted as successful only after the nut descends over the peg.

\subsection{Real-World Experiments}

The real-world experiment evaluates the PNN-LoRA framework, coordinated by the
LLM planner, on a long-horizon manipulation sequence using the FR3 arm (Fig.~\ref{fig:real_world_task}). The first two sub-tasks require full pose
control, including orientation, while the remaining four sub-tasks use position
control. The robot first picks up a plastic glass and empties its contents into a bowl,
then moves to a second glass and hangs it on a wall-mounted nail. It then opens a
box by displacing its lid, places a bowl and then a cup inside the box, and finally
restores the lid to close the box. Each stage is executed by its corresponding
expert policy.

Each expert is trained using $30$ demonstrations, and columns are added
incrementally so that each new skill can reuse previously learned frozen skills without causing forgetting. Figure~\ref{fig:box_close_real_world}(B) quantifies this reuse through the gates of the final box-close column. In this column, lateral paths account for $80\%$ of the measured transferred share, with $53\%$ coming from the five preceding real-world sub-skills and $27\%$ from the frozen simulation experts. The remaining $20\%$ is attributed to the new column's own path.

During execution, a sub-task is considered complete when the current object-pose error relative to its target falls below a predefined threshold. This signal is
returned to the LLM planner, which uses it to decide when to advance the plan and
which expert policy to invoke next. The sub-task order remains fixed across all
real-world trials. A trial is counted as successful only if every sub-task is
completed; under this criterion, the framework succeeds in four of five trials.

\section{Limitations \& Future Work}
The current framework does not use force or tactile feedback. Consequently, the hardware evaluation is limited to quasi-static manipulation and does not assess interactions that require compliant force regulation. Extending the action space to full pose control does not address this limitation because the policies still rely primarily on visual and kinematic state information. Another limitation is
that the current real-world evaluation uses a fixed sub-task order.

Future work will incorporate force and tactile feedback directly into the policy
input, allowing each expert to adapt its motion based on the measured contact
state. This extension could support force modulation during grasping,
insertion, sliding, and other contact-rich tasks. 
We will also evaluate more flexible long-horizon
instructions, including prompts with randomized sub-task order, to assess whether
the planner and expert library can complete all required sub-tasks reliably. Finally, we plan to investigate force-aware expert transfer, examining whether
previously learned contact strategies can be reused across tasks and embodiments
through the same gated low-rank lateral mechanism. 

\section{Conclusion}

We presented a continual robot learning framework that grows a library of
frozen PNN expert policies connected through gated LoRA lateral adapters. The
framework enables selective forward transfer without forgetting and is coordinated
by a frozen local LLM for long-horizon task decomposition and expert routing.
Across MetaWorld, cross-embodiment transfer, pose-control, and real-world
experiments on an FR3, the results show that reusable expert
libraries can reduce the data required to acquire new skills while supporting
compositional robot behavior.

{\footnotesize
\bibliographystyle{IEEEtran}
\bibliography{references}}

\end{document}